\documentclass[11pt,a4paper]{article}
\usepackage{times,latexsym}
\usepackage{url}
\usepackage[T1]{fontenc}

\usepackage[acceptedWithA]{tacl2018v2}

\usepackage{xspace,mfirstuc,tabulary}

\newif\iftaclinstructions
\taclinstructionsfalse 
\iftaclinstructions
\renewcommand{\confidential}{}
\renewcommand{\anonsubtext}{(No author info supplied here, for consistency with
TACL-submission anonymization requirements)}
\newcommand{\instr}
\fi

\iftaclpubformat 

\else

\fi

\usepackage{amsmath}
\usepackage{amssymb}
\usepackage{mathtools}
\usepackage{multirow}
\usepackage{multicol}
\usepackage{booktabs}
\usepackage{textcomp}
\usepackage{comment}
\usepackage{float}
\usepackage{ctable}
\usepackage{footmisc}
\usepackage{algorithm}
\usepackage[noend]{algpseudocode}
\usepackage[normalem]{ulem}

\usepackage{xcolor}
\definecolor{todored}{HTML}{F44336}
\definecolor{todogreen}{HTML}{008000}

\newcommand{\addcite}[1]{{\color{todored}~(cite?)}}

\newcommand{\secref}[1]{\textsection\ref{#1}}

\newcommand{\rev}[1]{#1}

\title{Deciphering Undersegmented Ancient Scripts Using Phonetic Prior}

\author{Jiaming Luo \\  CSAIL, MIT \\ {\sf j\_luo@csail.mit.edu}
        \And 
        Frederik Hartmann \\ University of Konstanz \\ {\sf frederik.hartmann@uni-konstanz.de} 
        \AND
        Enrico Santus \\ Bayer \\ {\sf enrico.santus@bayer.com
} \\ \\ 
        \And
        Yuan Cao \\ Google Brain \\ {\sf yuancao@google.com}
        \And
        Regina Barzilay \\ CSAIL, MIT \\ {\sf regina@csail.mit.edu}}

\date{}

\begin{document}
\maketitle

\begin{abstract}

Most undeciphered lost languages exhibit two characteristics that pose significant decipherment challenges: (1) the scripts are not fully segmented into words; (2) the closest known language is not determined. We propose a decipherment model that handles both of these challenges by building on rich linguistic constraints reflecting consistent patterns in historical sound change. We capture the natural phonological geometry by learning character embeddings based on the International Phonetic Alphabet (IPA).  The resulting generative framework jointly models word segmentation and cognate alignment, informed by phonological constraints. We evaluate the model on both deciphered languages (Gothic, Ugaritic) and an undeciphered one (Iberian).  The experiments show that incorporating phonetic geometry leads to clear and consistent gains.  Additionally, we propose a measure for language closeness which correctly identifies related languages for Gothic and Ugaritic. For Iberian, the method does not show strong evidence supporting Basque as a related language, concurring with the favored position by the current scholarship.\footnote{Code and data available at \url{https://github.com/j-luo93/DecipherUnsegmented/}.}
\end{abstract}

\section{Introduction}
All the known cases of lost language
decipherment have been accomplished by human experts, oftentimes over decades of painstaking efforts.  At least a dozen languages are still undeciphered today. For some of those languages, even the most fundamental questions pertaining to their origins and connections to known languages are shrouded in mystery, igniting fierce scientific debate among humanities scholars. Can NLP methods be helpful in bringing some clarity to these questions? Recent work has already demonstrated that algorithms can successfully decipher lost languages like Ugaritic and Linear B~\cite{luo2019neural}, relying only on non-parallel data in known languages -- Hebrew and Ancient Greek, respectively.  However, these methods are based on assumptions that are not applicable to many undeciphered scripts.


The first assumption relates to the knowledge of language family of the lost language. This information enables us to identify the closest living language, which anchors the decipherment process. Moreover, the models assume significant proximity between the two languages so that a significant portion of their vocabulary is matched. The second assumption presumes that word boundaries are provided, which uniquely defines the vocabulary of the lost language.

One of the famous counterexamples to both of these assumptions is Iberian.
The Iberian scripts are undersegmented with inconsistent use of word dividers. At the same time, there is no definitive consensus on its close known language  --- over the years, Greek, Latin and Basque were all considered as possibilities.

In this paper, we introduce a decipherment approach that relaxes the above assumptions. The model is provided with undersegmented inscriptions in the lost language and the vocabulary in a known language. No assumptions are made about the proximity between the lost and the known languages \rev{and the goal is to match spans in the lost texts with known tokens.} As a by-product of this model, we propose a measure of language closeness that drives the selection of the best target language from the wealth of world languages.

Given the vast space of possible mappings and the scarcity of guiding signal in the input data, decipherment algorithms are commonly informed by linguistic constraints. These constraints reflect consistent patterns in language change and linguistic borrowings. Examples of previously utilized constraints include skewness of vocabulary mapping, and monotonicity of character level alignment within cognates. We further expand the linguistic foundations of decipherment models, and incorporate phonological regularities of sound change into the matching procedure. For instance, a velar consonant [k] is unlikely to change into a labial [m]. Another important constraint in this class pertains to sound preservation, i.e., the size of phonological inventories is largely preserved during language evolution. 

Our approach is designed to encapsulate these constraints while addressing the segmentation issue. We devise a generative framework that jointly models word segmentation and cognate alignment. To capture the natural phonological geometry, we incorporate phonological features into character representations using the International Phonetic Alphabet (IPA). We introduce a regularization term to explicitly discourage the reduction of the phonological system and employ an edit distance-based formulation to model the monotonic alignment between cognates. The model is trained in an end-to-end fashion to optimize both the quality and the coverage of the matched tokens in the lost texts.

The ultimate goal of this work is to evaluate the model on an undeciphered language, specifically Iberian. Given how little is known about the language, it is impossible to directly assess prediction accuracy. Therefore, we adopt two complementary evaluation strategies to analyze model performance. First, we apply the model to deciphered ancient languages, Ugaritic and Gothic, which share some common challenges with Iberian.  Second, we consider evaluation scenarios which capitalize on a few known facts about Iberian, such as personal names, and report the model's accuracy against these ground truths. 

The results demonstrate that our model can robustly handle unsegmented or undersegmented scripts. In the Iberian personal name experiment, our model achieves a precision@10 score of 75.0\%.  Across all the evaluation scenarios, incorporating phonological geometry leads to clear and consistent gains. For instance, the model informed by IPA obtains 12.8\% increase in Gothic-Old Norse experiments. We also demonstrate that the proposed unsupervised measure of language closeness is consistent with historical linguistics findings on known languages.

\section{Related Work}

\paragraph{Non-parallel machine translation} 
At a high level, our work falls into research on non-parallel machine translation.
One of the important recent advancements in that area is the ability to induce accurate cross-lingual lexical representations without access to parallel data~\cite{lample2018word,lampleunsupervised,conneau2019cross}. This is achieved by aligning embedding spaces constructed from large amounts of monolingual data. The size of data for both languages is key: high-quality monolingual embeddings are required for successful matching.  This assumption, however, does not hold for ancient languages where we can typically access a few thousands of words at most.

\paragraph{Decoding cipher texts} Man-made ciphers have been the focal points for most of the early work on decipherment. They usually employ EM algorithms which are tailored towards these specific types of ciphers, most prominently substitution ciphers~\cite{knight.1999,knight-etal-2006-unsupervised}. Later work by ~\citet{nuhn2013beam,hauer2014solving,kambhatla2018decipherment} addresses the problem using a heuristic search procedure, guided by a pretrained language model. To the best of our knowledge, these methods developed for tackling man-made ciphers have so far not been successfully applied to archaeological data. One contributing factor could be the inherent complexity in the evolution of natural languages.

\begin{figure*}[t]
    \centering
    \includegraphics[width=0.83\linewidth]{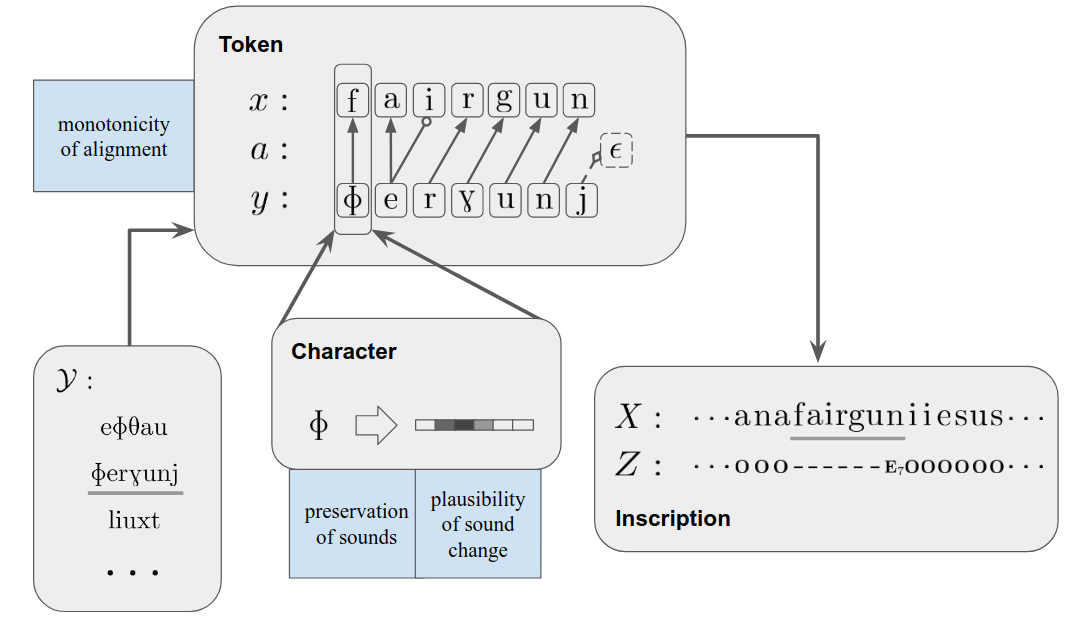}
    \caption{An overview of our framework which generates the lost texts from smaller units --- from characters to tokens and from tokens to inscriptions. Character mappings are first performed on the phonetic alphabet of the known language. Based on these mappings, a token $y$ in the known vocabulary $\mathcal{Y}$ is converted into a token $x$ in the lost language according to the latent alignment variable $a$. Lastly, all generated tokens, together with characters in unmatched spans, are concatenated to form a lost inscription. Blue boxes display the corresponding linguistic properties associated with each level of modeling.  }
    \label{fig:overview}
\end{figure*}

\paragraph{Deciphering ancient scripts} Our research is most closely aligned with computational decipherment of ancient scripts. Prior work has already featured several successful instances of ancient language decipherment previously done by human experts~\cite{snyder-etal-2010-statistical,kirkpatrick13decipherment,luo2019neural}. Our work incorporates many linguistic insights about the structure of valid alignments introduced in prior work, such as monotonicity. We further expand the linguistic foundation by incorporating phonetic regularities which have been beneficial in early, pre-neural decipherment work~\cite{knight-etal-2006-unsupervised}.  However, our model is designed to handle challenging cases not addressed by prior work, where segmentation of the ancient scripts is unknown. Moreover,  we are interested in dead languages without a known relative and introduce an unsupervised measure of language closeness that enables us to select an informative known language for decipherment.

\section{Model}
\newcommand{\X}{\mathcal{X}}
\newcommand{\Y}{\mathcal{Y}}
\newcommand{\Z}{\mathcal{Z}}
\newcommand{\Span}{\mathcal{S}}
\newcommand{\x}[1]{x^{#1}}
\newcommand{\xs}{\x{s}}
\newcommand{\xsi}{\x{s_i}}
\newcommand{\xsb}{\x{\bar{s}}}
\newcommand{\Xwt}{X[w_t]}
\newcommand{\set}[1]{\{#1\}}
\newcommand{\ispan}{matched span}
\newcommand{\Ispan}{Matched span}
\newcommand{\nispan}{unmatched span}
\newcommand{\lst}[2]{[#1_1, #1_2, \dots, #1_#2]}
\newcommand{\lost}{\mathbf{L}}
\newcommand{\known}{\mathbf{K}}
\newcommand{\IPA}{\textit{IPA}}
\newcommand{\other}{\textbf{O}}
\newcommand{\tagend}[1]{\textbf{E}_\textbf{#1}}
\newcommand{\pnis}{p_{0}}
\newcommand{\equationref}[1]{Equation~(\ref{#1})}
\newcommand{\score}{\texttt{score}}
\newcommand{\coverage}{\texttt{coverage}}
\newcommand{\zo}{\substack{z\in Z \\ z=\other}}
\newcommand{\znoto}{\substack{z\in Z \\ z\neq\other}}
\newcommand{\E}{\mathop{}{\mathbb{E}}}
\newcommand{\numberthis}{\addtocounter{equation}{1}\tag{\theequation}}
\newcommand{\cik}{c_i^{\known}}
\newcommand{\cjl}{c_j^{\lost}}
\newcommand{\nonctxp}{\Pr_{\text{nonctx}}}
\newcommand{\ctxp}{\Pr_{\text{ctx}}}
\newcommand{\nonctxpdot}{\Pr_{\text{nonctx}}(\cdot)}
\newcommand{\ctxpdot}{\Pr_{\text{ctx}}(\cdot)}

We design a model for the automatic extraction of  cognates\footnote{Throughout this paper, the term \emph{cognate} is liberally used to also include loanwords, as \rev{the sound correspondences in cognates and loanwords are both regular, although usually different.}} directly from unsegmented or undersegmented texts \rev{(detailed setting in Section~\ref{subsec:problem_setting})}. In order to properly handle the uncertainties caused by the issue of segmentation, we devise a generative framework which composes the lost texts using smaller units --- from characters to tokens, and from tokens to inscriptions. The model is trained in an end-to-end fashion to optimize both the quality and the coverage of the matched tokens.

To help the model navigate the complex search space, we consider the following linguistic properties of sound change, including phonology and phonetics in our model design:
\begin{itemize}
    \item \textbf{Plausibility of sound change}: Similar sounds rarely change into drastically different sounds. This pattern is captured by the natural phonological geometry in human speech sounds and we incorporate relevant phonological features into the representation of characters.
    \item \textbf{Preservation of sounds}: The size of phonological inventories tends to be largely preserved over time. This implies that total disappearance of any sound is uncommon. In light of this, we employ a regularization term to discourage any sound loss in the phonological system of the lost language.
    \item \textbf{Monotonicity of alignment}: The alignment between any matched pair is predominantly monotonic\rev{, which means that character-level alignments do not cross each other.} This property inspires our edit distance-based formulation at the token level.
\end{itemize}

To reason about phonetic proximity, we need to find character representation that explicitly reflects its phonetic properties. One such representation is provided by the International Phonetic Alphabet (IPA), where each character is represented by a vector of phonological features. As an example, consider IPA representation 
for two phonetically close characters [b] and [p] (See Figure~\ref{fig:IPA}), which share two identical coordinates. To further refine this representation, the model learns to embed these features into a new space, optimized for the decipherment task.


\subsection{Problem setting}
\label{subsec:problem_setting}
We are given a list of \emph{unsegmented} or \emph{undersegmented} inscriptions $\X = \set{X}$ in the lost language, and a vocabulary, i.e., a list of tokens $\Y = \set{y}$ in the known language. For each lost text $X$, the goal is to identify a list of non-overlapping spans $\set{x}$ that correspond to cognates in $\Y$. We refer to these spans as \textbf{\ispan{s}} and any remaining character as \textbf{\nispan{s}}.

We denote the character sets of the lost and the known languages by $C^\lost =\set{c^{\lost}}$ and $C^\known = \set{c^{\known}}$, respectively.
To exploit the phonetic prior, IPA transcriptions are used for $C^\known$, while orthographic characters are used for $C^\lost$. \rev{For this paper, we only consider alphabetical scripts for the lost language.}\footnote{Given that the known side uses IPA, an alphabetical system, having an alphabetical system on the lost side makes it much easier to enforce the linguistic constraints in this paper. For other types of scripts, it requires more thorough investigation which is beyond the scope of this work.}

\subsection{Generative framework}
\label{subsec:generative_framework}

\begin{figure}
    \centering
    \includegraphics[width=0.7\linewidth]{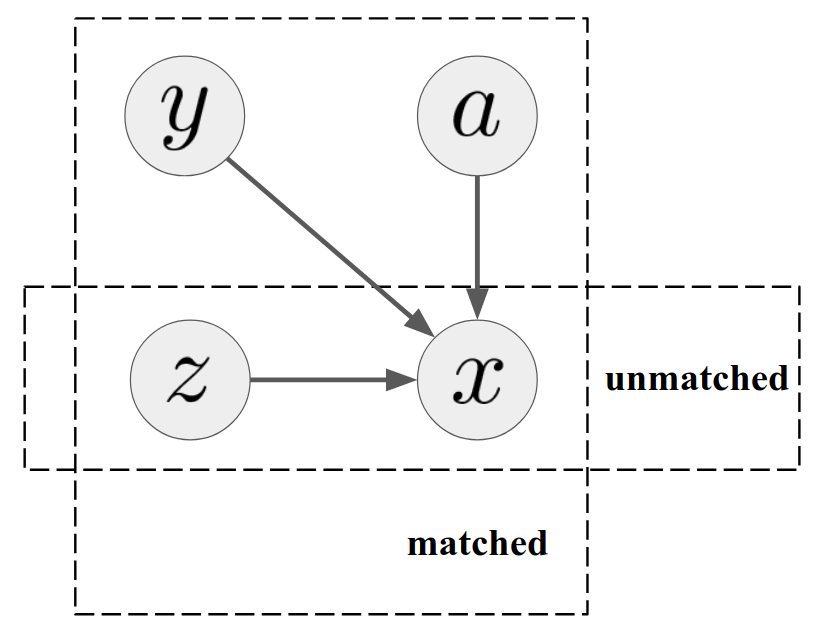}
    \caption{A graphical model representation for our framework to generate a span $x$. Characters in unmatched spans are generated in an i.i.d. fashion whereas matched spans are additionally conditioned on two latent variables: $y$ representing a known cognate and $a$ the character-level alignment between $x$ and $y$.}
    \label{fig:graphical}
\end{figure}

We design the following generative framework  to  handle the issue of segmentation. It jointly models segmentation and cognate alignment, which requires different treatments for \ispan{s} and \nispan{s}. An overview of the framework is provided in Figure~\ref{fig:overview} and  a graphical model representation in Figure~\ref{fig:graphical}.

For \ispan{s}, we introduce two latent variables: $y$ representing the corresponding cognate in the known language and $a$ indicating the character alignment between $x$ and $y$ (see the \textit{Token} box in Figure~\ref{fig:overview}). More concretely, $a=\set{a_\tau}$ is a sequence of indices, with $a_\tau$ representing the aligned position for $y_\tau$ in $x$. The lost token is generated by applying a character-level mapping to $y$ according to the alignment provided by $a$.  For \nispan{s}, we assume each character is generated in an i.i.d. fashion under a uniform distribution $\pnis = \frac{1}{|C^\lost|}$.

Whether a span is matched or not is indicated by another latent variable $z$, and the corresponding span is denoted by $x_z$. More specifically, each character in an \nispan{} is tagged by $z=\other$ whereas the entirety of a \ispan{} of length $l$ is marked by $z=\tagend{l}$ at the end of the span (see the \emph{Inscription} box in Figure~\ref{fig:overview}). All spans are then concatenated to form the inscription, with a corresponding (sparse) tag sequence $Z=\set{z}$.

Under this framework, we arrive at the following derivation for the marginal distribution for each lost inscription $X$:
\begin{alignat}{3}
    &&&\Pr(X)  \nonumber \\
    &=&& \sum_{Z}\Big[\prod_{z\in Z} \Pr(z)\Big] \Big[\prod_{\zo}\pnis\Big] \Big[\prod_{\znoto}\Pr(x_z|z)\Big]
    \label{eq:marginal_px}
\end{alignat}
where $\Pr(x_z | z\neq\other)$ is further broken down into individual character mappings:
\begin{alignat}{3}
     &&&\Pr(x_z | z\neq\other)\nonumber \\
     &=&&  \sum_{y\in\Y} \sum_{a\in\mathcal{A}} \Pr(y) \Pr(a) \cdot\Pr(x_z |y, z, a) \nonumber\\
 &\propto&& \sum_{y\in\Y} \sum_{a\in\mathcal{A}} \Pr(x_z | y, z, a)  \nonumber \\
 &\approx&& \sum_{y\in\Y} \max_a \Pr(x_z | y, z, a)  \nonumber \\
 &=&& \sum_{y\in\Y} \max_a \prod_\tau \Pr(x_{a_\tau} | y_\tau) \label{eq:p_x_z}
\end{alignat}
Note that we assume a uniform prior for both $y$ and $a$, and use the maximum to approximate the sum of $\Pr(x_z | y, z, a)$ over the latent variable $a$. $\mathcal{A}$ is the set of valid alignment values to be detailed in~\secref{subsubsec:edit_dist}.

\subsubsection{Phonetics-aware parameterization}
\label{subsubsec:parameterization}

\begin{figure}
    \centering
    \includegraphics[width=0.9\linewidth]{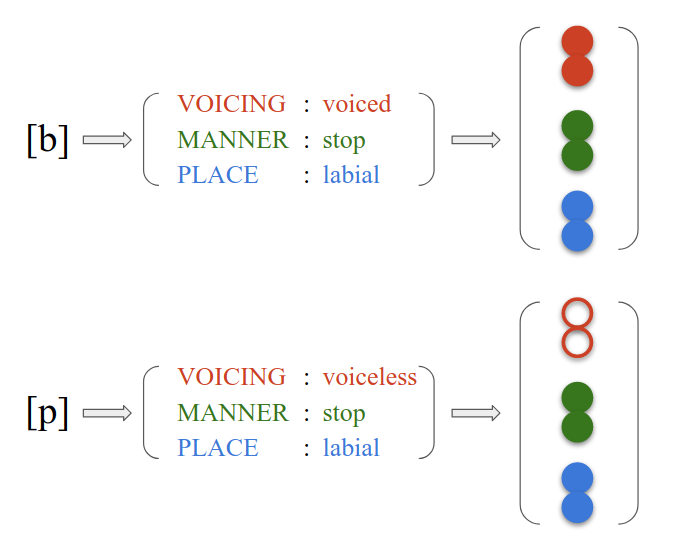}
    \caption{An illustration of IPA embeddings. Each phone is first represented by a vector of phonological features. The model first embeds each feature and then IPA embedding is obtained by concatenating all its relevant feature embeddings. For instance, the phone [b] can be represented as  the concatenation of the \texttt{voiced}, \texttt{stop} and the \texttt{labial} embeddings.}
    \label{fig:IPA}
\end{figure}

The character mapping distributions are specified as follows:
\begin{alignat}{3}
   &&&\Pr\nolimits_\theta(x_{a_\tau} = \cjl| y_\tau = \cik) \nonumber \\
   &\propto&& \exp\bigg(\frac{E^{\lost}(\cjl) \cdot E^{\known}(\cik)}{T}   \bigg), 
\end{alignat}
where $T$ is a temperature hyperparameter,  $E^\lost(\cdot)$ and $E^\known(\cdot)$ are the embedding functions for the lost characters and the known characters, respectively, and $\theta$ is the collection of all trainable parameters (i.e., the embeddings).

In order to capture the similarity within certain sound classes, we use IPA embeddings to represent each IPA character in the known language. More specifically, each IPA character is represented by a vector of phonological features. 
The model learns to embed these features into a new space and the full IPA embedding for $c^\known$ is composed by concatenating all of its relevant feature embeddings. For the example in Figure~\ref{fig:IPA}, the phone [b] can be represented as  the concatenation of the \texttt{voiced} embedding, the \texttt{stop} embedding and the \texttt{labial} embedding.

This compositional structure encodes the natural geometry existent in sound classes~\cite{stevens2000acoustic} and biases the model towards utilizing such a structure. By design, the representations for [b] and [p] are close as they share the same values for two out of three feature groups. This structural bias is crucial for realistic character mappings.

For the lost language, we represent each character $\cjl$ as a weighted sum of IPA embeddings on the known side. Specifically,
\begin{align}
    E^{\lost}(\cjl) = \sum_i w_{i, j} \cdot E^{\known}(\cik),
\end{align}
where $\{w_{i,j}\}$  are learnable parameters.

\subsubsection{Monotonic alignment and edit distance}
\label{subsubsec:edit_dist}
Individual characters in the known token $y$ are mapped to a lost token $x$ according to the alignment variable $a$. The monotonic nature of character alignment between cognate pairings motivates our design of an edit distance-based formulation to capture the dominant mechanisms involved in cognate pairings: substitutions, deletions and insertions~\cite{campbell2013historical}. In addition to $a_\tau$ taking the value of an integer signifying the substituted position, $a_\tau$ can be $\epsilon$, which indicates that $y_\tau$ is deleted. To model insertions, $a_\tau=(a_{\tau, 1}, a_{\tau, 2})$ can be two\footnote{Insertions of even longer character sequences are rare.} adjacent indices in $x$.

This formulation inherently defines a set $\mathcal{A}$  of valid values  for the alignment. Firstly, they are monotonically increasing with respect to $\tau$, with the exception of $\epsilon$. Secondly, they cover every index of $x$, which means every character in $x$ is accounted for by some character in $y$. The \emph{Token} box in Figure~\ref{fig:overview} showcases such an example with all three types of edit operations.  More concretely, we have the following alignment model:
\begin{align}
    \Pr(x_{a_\tau} | y_{\tau}) &= 
    \begin{cases}
        \Pr_\theta(x_{a_\tau} | y_{\tau}) & \text{(substitution)}\\
        \Pr_\theta(\epsilon | y_{\tau}) & \text{(deletion)} \\
        \Pr_\theta(x_{a_{\tau, 1}} | y_\tau) \nonumber \\
    \cdot\alpha \Pr_\theta(x_{a_{\tau, 2}} | y_\tau)\quad & \text{(insertion)}
    \end{cases}
\end{align}
where $\alpha \in [0, 1]$ is a hyperparameter to control the use of insertions. 

\subsection{Objective}
\label{subsec:objective}
Given the generative framework, our training objective is designed to optimize the quality of the extracted cognates, while matching a reasonable proportion of the text.

\begin{algorithm*}[th!]
 \caption{One training step for our decipherment model}
  \label{alg}
{\normalsize  
	\textbf{Input:}  One batch of lost inscriptions $\tilde{\X}$, entire known vocabulary $Y=\set{y}$ \\
	\textbf{Parameters:} Feature embeddings $\theta$
	\begin{algorithmic}[1]
        \State $\Pr(\cjl | \cik) \gets \textit{ComputeCharDistr}(\theta)$ \Comment{Compute character mapping distributions (Section~\ref{subsubsec:parameterization})}
    \State $\Pr(x | y) \gets \textit{EditDistDP}\big(x, y, \Pr(\cjl | \cik)\big)$ \Comment{Compute token alignment probability (Section~\ref{subsubsec:edit_dist})}
    \State $\mathcal{S}(\tilde{\X}; C^\lost, C^\known) \gets \textit{WordBoundaryDP}\big(\Pr(x | y)\big)$ \Comment{Compute final objective (Section~\ref{subsec:objective})}
    \State $\theta \gets \textit{SGD}(\mathcal{S})$ \Comment{Backprop and update parameters}

	\end{algorithmic}
        }
\end{algorithm*}

\paragraph{Quality}
We aim to optimize the \textit{quality} of \ispan{s} under the posterior distribution $\Pr(Z|X)$, measured by a scoring function $\Phi(X, Z)$. $\Phi(X, Z)$ is computed by  aggregating the likelihoods of these \ispan{s} normalized by length. The objective is defined  as follows:
\begin{align}
    Q(X) &= \E_{Z\sim \Pr(Z | X)} \Phi(X, Z) \label{eq:quality} \\ 
    \Phi(X, Z) &= \sum_{\znoto} \phi(x_z, z) \\ 
    \phi(x_z, z) &= \Pr(x_z | z)^\frac{1}{|x_z|}
\end{align}
This term encourages the model to explicitly focus on improving the probability of generating the \ispan{s}.

\paragraph{Regularity and coverage}
The regularity of sound change, as stated by the Neogrammarian hypothesis~\cite{campbell2013historical}, implies that we need to find a reasonable number of matched spans. To achieve this goal, we incur a penalty if the expected coverage ratio of the matched characters under the posterior distribution falls below a given threshold $r_{\text{cov}}$:
\begin{align}
    \Omega_{\text{cov}}(\X) &= \max\Big(r_{\text{cov}}-\frac{\sum_{X\in\X}  \texttt{cov}(X)}{|\X|}, 0.0\Big) \nonumber \\
    \texttt{cov}(X) &= \E_{Z\sim\Pr(Z | X)} \Psi(X, Z) \label{eq:coverage} \\
    \Psi(X, Z) &= \sum_{\znoto} \psi(x_z, z) = \sum_{\znoto} |x_z|
\end{align}
Note that the ratio is computed on the entire corpus $\X$ instead of individual texts $X$ since the coverage ratio can vary greatly for different individual texts. The hyperparameter $r_{\text{cov}}$ controls the expected overlap between two languages, which enables us to apply the method even when languages share some loanwords but are not closely related.

\paragraph{Preservation of sounds}
The size of phonological inventories tends to be largely preserved over time.
This implies that total disappearance of any sound is uncommon. To reflect this tendency, 
we introduce an additional regularization term to discourage any sound loss. The intuition is to encourage any lost character to be mapped to exactly one\footnote{We experimented with looser constraints (e.g., with \emph{at least} instead of \emph{exactly} one correspondence), but obtained worse results.} known IPA symbol. Formally we have the following term
\begin{align*}
    \Omega_{\text{loss}}(C^\lost, C^\known) &= \sum_{c^\lost} \big(\sum_{c^\known} \Pr(c^\lost | c^\known) - 1.0\big) ^ 2
\end{align*}

\paragraph{Final objective}
Putting the terms together, we have the following final objective:
\begin{alignat}{3}
    \mathcal{S}(\X; C^\lost, C^\known)\nonumber 
    = &\sum_{X\in\X} Q(X)
    + \lambda_{\text{cov}}\Omega_{\text{cov}}(\X) \nonumber\\
    &+ \lambda_{\text{loss}} \Omega_{\text{loss}}(C^\lost, C^\known)
\end{alignat}
where $\lambda_{\text{cov}}$ and  $\lambda_{\text{loss}}$ are both hyperparameters.

\subsection{Training}
\label{subsec:training}
Training with the final objective involves either finding the best latent variable, as in~\equationref{eq:p_x_z}, or computing the expectation under a distribution that involves one latent variable, as in~\equationref{eq:quality} and~\equationref{eq:coverage}.
In both cases, we resort to dynamic programming to facilitate efficient computation and end-to-end training. We refer interested readers to~\ref{app:dp} for more detailed derivations. \rev{We illustrate one training step in Algorithm~\ref{alg}.}

\section{Experimental setup}

\ctable[
caption={Basic information about the lost languages},
label=tab:lost,
pos=thb,
star,
botcap
]{lllrlr}{
\tnote[$\dagger$]{\url{http://www.wulfila.be/gothic/download/}}
\tnote[$\dagger\dagger$]{\url{http://hesperia.ucm.es/}. \rev{Iberian language is semi-syllabic, but this database has already transliterated the inscriptions into Latin scripts.}}
\tnote[$\ddagger$]{This dataset directly provides the Ugaritic vocabulary, i.e., each word occurs exactly once.}
\tnote[$\ddagger\ddagger$]{Since the texts are undersegmented and we do not know the ground truth segmentations, this represents the number of unsegmented \emph{chunks}, each of which might contain multiple tokens.}
}{ 
\multicolumn{1}{l}{\textbf{Language}} & \textbf{Family}   & \multicolumn{1}{l}{\textbf{Source}} & \multicolumn{1}{c}{\textbf{\#Tokens}} & \begin{tabular}[c]{@{}c@{}}\textbf{Segmentation}\\ \textbf{situation}\end{tabular} & \multicolumn{1}{c}{\textbf{Century}} \\
\FL
Gothic & Germanic & Wulfila\tmark[$\dagger$]& 40,518 & unsegmented    & 3–10 AD\\
Ugaritic & Semitic  &  \citet{snyder-etal-2010-statistical} & 7,353\rlap{\tmark[$\ddagger$]} & segmented & 14–12 BC\\
Iberian & unclassified  & Hesperia\tmark[$\dagger\dagger$]    & 3,466\rlap{\tmark[$\ddagger\ddagger$]} & undersegmented & 6–1 BC
\LL
}

Our ultimate goal is to evaluate the decipherment capacity for unsegmented lost languages, without information about a known counterpart. Iberian fits both of these criteria. However, our ability to evaluate decipherment of Iberian is limited since a full ground truth is not known. Therefore, we supplement our evaluation on Iberian with more complete evaluation on lost languages with known translation, such as Gothic and Ugaritic.

\subsection{Languages}
 We focus our description on the Gothic and Iberian corpora which we compiled for this paper. Ugaritic data was reused from the prior work on decipherment~\cite{snyder-etal-2010-statistical}. Table~\ref{tab:lost} provides statistics about these languages. To evaluate the validity for our proposed language proximity measure, we additionally include six known languages: Spanish (Romance), Arabic (Semitic), Hungarian (Uralic), Turkish (Turkic), classical Latin (Latino-Faliscan) and Basque (isolate).  

\paragraph{Gothic} Several features of Gothic make it an ideal candidate for studying decipherment models. Since Gothic is fully deciphered, we can compare our predictions against ground truth. Like Iberian, Gothic is unsegmented. Its alphabet was adapted from a diverse set of languages: Greek, Latin and Runic, but some characters are of unknown origin. The latter were in the center of decipherment efforts on Gothic~\cite{zacher.1855,wagner.2006}. Another appealing feature of Gothic is its relatedness to several known Germanic languages which exhibit various degree of proximity to Gothic. The closest is its reconstructed ancestor Proto-Germanic, with Old Norse and Old English being more distantly related to Gothic. This variation in linguistic proximity enables us to study the robustness of decipherment methods to the historical change in the source and the target.

\paragraph{Iberian} Iberian serves as a real test scenario for automatic methods -- it is still undeciphered, withstanding multiple attempts over at least two centuries. Iberian scripts present two issues facing many undeciphered languages today: undersegmentation and lack of a well-researched relative. Many theories of origin have been proposed in the past, most notably linking Iberian to Basque, another non-Indo-European language on the Iberian peninsular. However, due to a lack of conclusive evidence, the current scholarship favors the position that Iberian is not genetically related to any living language. Our knowledge of Iberian owes much to the phonological system proposed by Manuel Gómez Moreno in the mid 20th century, based on fragmentary evidences such as bilingual coin legends~\cite{sinner2019palaeohispanic}. Another area with a broad consensus relates to Iberian personal names, thanks to a key Latin epigraph, \textit{Ascoli Bronze}, which recorded the grant of Roman citizenship to Iberian soldiers who had fought for Rome~\cite{marti2017indigenous}. We use these personal names recorded in Latin as the known vocabulary.

\subsection{Evaluation}

\paragraph{Stemming and Segmentation} Our matching process operates at the \emph{stem} level for the known language, instead of \emph{full words}. Stems are more consistently preserved during 
language change or linguistic borrowings. While we always assume that gold stems 
are provided for the known language, we estimate them for the lost language.

The original Gothic texts are only segmented into sentences. To study the effect of having varying degrees of prior knowledge about the word segmentations, we create separate datasets by randomly inserting ground truth segmentations (i.e., whitespaces) with a preset probability to simulate undersegmentation scenarios.

\ctable[
caption={Main results on Gothic in a variety of settings using P@10 scores. All scores are reported in the format of triplets, corresponding to \texttt{base} / \texttt{partial} / \texttt{full} models. In general, more phonological knowledge about the lost language, more segmentations improve the model performance. The choice of the known language also plays a significant role as Proto-Germanic has a noticeably higher score than the other two choices.},
label={tab:gothic_main_results_compact},
pos=ht,
star,
botcap]{ccccc}{
\tnote[$\dagger$]{Short for \emph{whitespace ratio}.}
\tnote[$\ddagger$]{Averaged over all whitespace ratio values.}
\tnote[$\dagger\dagger$]{Averaged over all known languages.}
}{
\centering
\multirow{2}{*}{\textbf{WR}\tmark[$\dagger$]} & \multicolumn{4}{c}{\textbf{Known language}} \\ 
\cline{2-5}\cline{2-5}
& {Proto-Germanic} (PG) & {Old Norse} (ON) & {Old English} (OE) & avg\tmark[$\dagger\dagger$] \\
\toprule
0\% & 0.820 / 0.749 / 0.863 &	0.213 / 0.397 / 0.597 &	0.046 / 0.204 / 0.497 & 0.360 / 0.450 / 0.652  \\
25\% & 0.752 / 0.734 / 0.826 &	0.312 / 0.478 / 0.610 &	0.128 / 0.328 / 0.474 & 0.398 / 0.513 / 0.637  \\
50\% &0.752 / 0.736 / 0.848 & 0.391 / 0.508 / 0.643	&0.169 / 0.404 / 0.495 & 0.438 / 0.549 / 0.662 \\
75\% & 0.761 / 0.732 / 0.866 &	0.435 / 0.544 / 0.682	&0.250 / 0.447 / 0.533 & 0.482 / 0.574 / 0.693 \\
\midrule
avg\tmark[$\ddagger$] & 0.771 / 0.737 / 0.851 & 0.338 / 0.482 / 0.633 & 0.148 / 0.346 / 0.500 & 0.419 / 0.522 / 0.661 \LL
}

\paragraph{Model Variants}
In multiple decipherment scenarios, partial information about phonetic assignments is available. This is the case with both Iberian and Gothic. Therefore, we evaluate performance of our model with respect to available  phonological knowledge for the lost language. The \texttt{base} model assumes no knowledge while the \texttt{full} model has full knowledge of the phonological system and therefore the character mappings. For the Gothic experiment, we additionally experiment with a \texttt{partial} model which assumes that we know the phonetic values for the characters \textit{k}, \textit{l}, \textit{m}, \textit{n}, \textit{p}, \textit{s}  and \textit{t}. The sound values of these characters can be used as prior knowledge as they closely resemble their original counterparts in Latin or Greek alphabets.
These known mappings are incorporated through an additional term which encourages the model to match its predicted distributions with the ground truths.

In scenarios with full segmentations where it is possible to compare with previous work, we report the results for the \texttt{Bayesian}  model  proposed by~\citet{snyder-etal-2010-statistical} and \texttt{NeuroCipher} by~\citet{luo2019neural}.

\paragraph{Metric} We evaluate the model performance using precision at K (P@K) scores. The prediction (i.e., the stem-span pair) is considered correct if and only if the stem is correct and the the span is the prefix of the ground truth. \rev{For instance, the ground truth for the Gothic word \textit{garda} has the stem \textit{gard} spanning the first four letters, matching the Old Norse stem \textit{gar\dh}. We only consider the prediction as correct if it correctly matches \textit{gar\dh} and the predicted span starts with the first letter.}

\section{Results}

\ctable[
caption={Results for comparing \texttt{base} model with previous work. \texttt{Bayesian} and \texttt{NeuroCipher} are the models proposed by~\citet{snyder-etal-2010-statistical} and~\citet{luo2019neural}, respectively. Ugaritic results for previous work are taken from their papers. For \texttt{NeuroCipher}, we run the authors' public implementation to obtain the results for Gothic.},
label=tab:comparison,
pos=h,
star,
botcap]{@{\extracolsep{15pt}}lcccc@{}}
{
\tnote[$\dagger$]{P@1 is reported for Ugaritic to make direct comparison with previous work. P@10 is still used for Gothic experiments.}
}{
\textbf{Lost}  & Ugaritic\tmark[$\dagger$] & \multicolumn{3}{c}{Gothic}                          \ML
\textbf{Known} & Hebrew   & \multicolumn{1}{c}{PG} & \multicolumn{1}{c}{ON} & OE    \\  \cline{2-2}\cline{3-5} \midrule
\texttt{Bayesian} & 0.604    & - & - & -     \\
\texttt{NeuroCipher} & 0.659    & 0.753 & 0.543 & 0.313 \\
\texttt{base} & \textbf{0.778}    & \textbf{0.865} & \textbf{0.558} & \textbf{0.472} \LL
}

\begin{figure*}[h]
    \centering
    \includegraphics[width=0.8\linewidth]{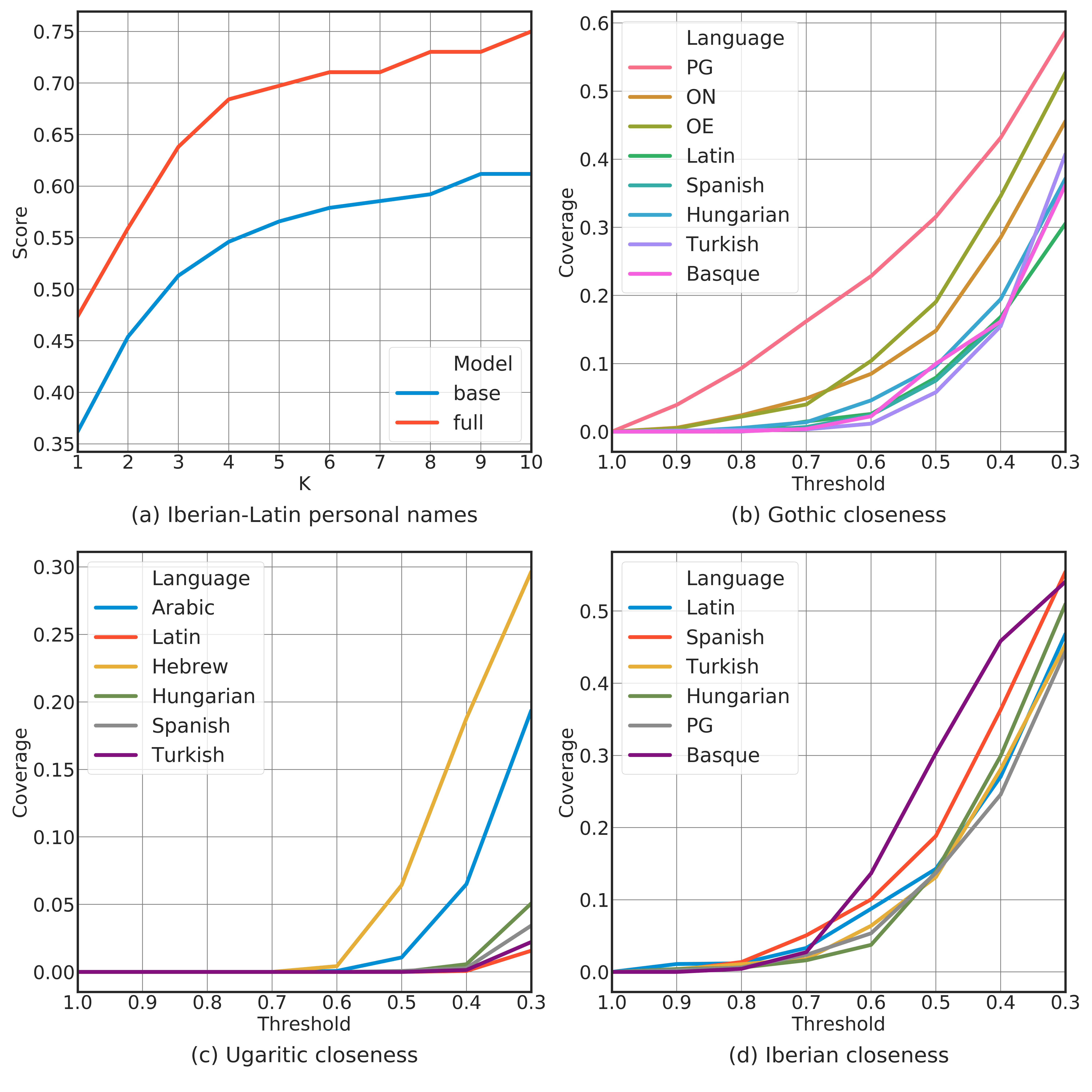}
    \caption{(a) P@K scores on Iberian using personal name recorded in Latin; (b), (c) and (d): Closeness plots for Gothic, Ugaritic and Iberian, respectively. }
    \label{fig:ono_closeness}
\end{figure*}

\paragraph{Decipherment of undersegmentated texts}
Our main results on Gothic in Table~\ref{tab:gothic_main_results_compact} demonstrate that our model can effectively extract cognates in a variety of settings. Averaged over all choices of whitespace ratios and known languages (bottom right), our \texttt{base}/\texttt{partial}/\texttt{full} models achieve P@10 scores of 0.419/0.522/0.661, respectively.  Not surprisingly, access to additional knowledge either about phonological mappings and/or segmentation lead to improved performance.

The choice of the known language also plays a significant role. 
On the closest language pair Gothic-PG, P@10 reaches 75\% even without assuming any phonological knowledge about the lost language. As expected, language proximity directly impacts the complexity of the decipherment tasks which in turn translates into lower model performance on Old English and Old Norse. These results reaffirm that choosing a close known language is vital for decipherment.

The results on Iberian shows that our model performs well on a real undeciphered language with undersegmented texts. As shown in Figure~\ref{fig:ono_closeness}a, \texttt{base} model reaches 60\% in P@10 while \texttt{full} model reaches 75\%. Note that Iberian is non-Indo-European with no genetic relationship with Latin,  but our model can still discover regular correspondences for this particular set of personal names.

\paragraph{Ablation study} To investigate the contribution of  phonetic and phonological knowledge, we conduct an ablation study using Gothic/Old Norse (Table~\ref{tab:ablation}). The IPA embeddings consistently improve all the model variants. As expected, the gains are most noticeable -- +12.8\% -- for the hardest matching scenario where no prior information is available (\texttt{base} model). As expected, $\Omega_{\text{loss}}$ is vital for \texttt{base} but unnecessary for \texttt{full} which 
has readily available character mapping.

\begin{table}
\centering
\begin{tabular}{ccccc}
IPA & $\Omega_{\text{loss}}$ & \texttt{base} & \texttt{partial} & \texttt{full} \\ \FL
+ & + & 0.435 & 0.544 & 0.682 \\
- & + & 0.307 & 0.490 & 0.599 \\
+ & - & 0.000 & 0.493 & 0.695
\end{tabular}
\caption{Ablation study on the pair Gothic-ON. Both IPA embeddings and the regularization on sound loss are beneficial, especially when we do not assume much phonological knowledge about the lost language.}
\label{tab:ablation}
\end{table}

\paragraph{Comparison with previous work}  To compare with the
state-of-the-art decipherment models~\cite{snyder-etal-2010-statistical,luo2019neural}, we consider the version of our model that operates with 100\% whitespace ratio for the lost language. Table~\ref{tab:comparison} demonstrates that our model consistently outperforms the baselines for both Ugaritic and
Gothic. For instance, it reaches over 11\% gain for Hebrew/Ugaritic pair and over 15\% for Gotchic/Old English.

\paragraph{Identifying close known languages}
Next we evaluate model's ability to identify a close known language to anchor the decipherment process. We expect that for a closer language pair, the predictions of the model will be more confident while matching more characters. We illustrate this idea with a plot that charts \textit{character coverage} (i.e., what percentage of the lost texts are matched regardless of its correctness) as a function of \textit{prediction confidence} value (i.e., probability of generating this span normalized by its length).  As Figure~\ref{fig:ono_closeness}b and Figure~\ref{fig:ono_closeness}c illustrate the model accurately predicts the closest languages for both Ugaritic and Gothic. Moreover, languages within the same family as the lost language stand out from the rest.

The picture is quite different for Iberian (see Figure~\ref{fig:ono_closeness}d). No language seems to have a pronounced advantage over others. This seems to accord with the current scholarly understanding that Iberian is a language isolate, with no established kinship with others. Basque somewhat stands out from the rest, which might be attributed to its similar phonological system with Iberian~\cite{sinner2019palaeohispanic} and very limited
vocabulary overlap (numeral names)~\cite{aznar2005algunos} which doesn't carry over to the lexical system.\footnote{\rev{For true isolates, whether the predicted segmentations are reliable despite the lack of cognates is beyond our current scope of investigation.} }

\begin{table}[]
    \centering
    \begin{tabular}{ll}
    \textbf{Inscription} & \textbf{Matched stem} \\ 
    \toprule    
    \th ammuhsamin\textcolor{blue}{haid}au & \textcolor{blue}{xai\dh} \\
    \midrule
    \th ammuhsamin{\color{todogreen}haid}au & {\color{todogreen}xai\dh} \\
     \th ammuhsamin{\color{todogreen}haid}au & {\color{todored}rai\dh} \\
    \th ammuhsamin{\color{todogreen}haid}au & {\color{todored}brai\dh} \\
    \end{tabular}
    \caption{\rev{One example of top 3 model predictions for \texttt{base} on Gothic-PG in WR 0\% setting. Spans are highlighted in the inscriptions. The first row presents the ground truth and the others are the model predictions. Green color is used for correct predictions and red for incorrect ones.}}
    \label{tab:example}
\end{table}

\section{Conclusions}
We propose a decipherment model to extract cognates from undersegmented texts, without assuming proximity between lost and known languages. Linguistics properties are incorporated into the model design, such as phonetic plausibility of sound change and preservation of sounds. Our results on Gothic, Ugaritic and Iberian shows that our model can effectively handle undersegmented texts even when source and target languages are not related. Additionally, we introduce a method for identifying close languages which correctly finds related languages for Gothic and Ugaritic. For Iberian, the method does not show strong evidence supporting Basque as a related language, concurring with the favored position by current scholarship.

\rev{Potential applications of our method are not limited to decipherment.
The phonetic values of lost characters can be inferred by mapping them to the known cognates. These values can serve as the starting point for lost sound reconstruction and more investigation is needed to establish their efficacy. Moreover, the effectiveness of incorporating phonological feature embeddings provides a path for future improvement for cognate detection in computational historical linguistics~\cite{rama2019automated}. Currently our method operates on a pair of languages. To simultaneously process multiple languages as it is common in the cognate detection task, more work is needed to modify our current model and its inference procedure.}

\section*{Acknowledgments}
We sincerely thank Noemí Moncunill Martí for her invaluable guidance on Iberian onomastics, and  Eduardo Orduña Aznar for his tremendous help on the Hesperia database and the Vasco-Iberian theories. Special thanks also go to Ignacio Fuentes and Carme  Huertas for the insightful discussions. 
This research is based upon work supported in part by the Office of the Director of National Intelligence (ODNI), Intelligence Advanced Research Projects Activity (IARPA), via contract \# FA8650-17-C-9116. The views and conclusions contained herein are those of the authors and should not be interpreted as necessarily representing the official policies, either expressed or implied, of ODNI, IARPA, or the U.S. Government. The U.S. Government is authorized to reproduce and distribute reprints for governmental purposes notwithstanding any copyright annotation therein.

\bibliography{anthology,acl2020,tacl2018}
\bibliographystyle{acl_natbib}

\clearpage
\appendix
\section{Appendices}
\subsection{Derivations for dynamic programming}
\label{app:dp}
We show the derivation for $\Pr(X)$ here --- other quantities can be derived in a similar fashion.

Given any $X$ with length $n$, let $p_{i}(X)$ be the probability of generating the prefix subsequence $X_{:i}$, and $p_{i,z}(X)$ be the probability of generating $X_{:i}$ using $z$ as the \emph{last} latent variable. By definition, we have
\begin{align}
    \Pr(X) &= p_{n}(X) \\
    p_{i}(X) &= \sum_z p_{i, z}(X)
\end{align} $p_{i, z}$ can be recursively computed using the following equations:
\begin{alignat}{3}
    p_{i, \other} &=&&\ \Pr(\other) \cdot \pnis \cdot p_{i - 1} \label{eq:pio} \\
    p_{i, \tagend{l}} &=&&\ \Pr(\tagend{l}) \cdot \Pr(x_{i-l+1:l}|\tagend{l})\cdot p_{i - l} \label{eq:recursive_word}
\end{alignat}

\subsection{Data preparation}
\paragraph{Stemming} 
Gothic stemmers are developed based on the documentations of \texttt{Gomorphv2}\footnote{\url{http://www.wulfila.be/gomorph/gothic/html/}}. Stemmers for Proto-Germanic, Old Norse and Old English are derived from relevant Wikipedia entries on their grammar and phonology. For all other languages, we use the Snowball stemmer from NLTK~\cite{bird-2006-nltk}. 

\paragraph{IPA transcription}
We use the \texttt{CLTK} library\footnote{\url{http://cltk.org/}} for Old Norse and Old English, and a rule-based converter for Proto-Germanic based on \cite[][pp. 242-260]{Ringe.2017}. Basque transcriber is based on its Wikipedia guide for transcription, and all other languages are transcribed using Epitran~\cite{Mortensen-et-al:2018}. The \texttt{ipapy} library\footnote{\url{https://github.com/pettarin/ipapy}} is used to obtain their phonetic features. There are 7 feature groups in total.

\paragraph{Known vocabulary}
For Proto-Germanic, Old Norse and Old English, we extract the information from the descendant trees in Wiktionary\footnote{\url{https://www.wiktionary.org/}}. All matched stems with at least four characters form the known vocabulary. It resulted in 7883, 10754 and 11067 matches with Gothic inscriptions, and 613, 529, 627 unique words in the vocabularies for Proto-Germanic, Old Norse and Old English, respectively. For Ugaritic-Hebrew, we retain stems with at least three characters due to its shorter average stem length. For the Iberian-Latin personal name experiments, we take the list provided by~\citet{ramos2014nuevo} and select the elements that have both Latin and Iberian correspondences. We obtain 64 unique Latin stems in total.
For Basque, we use a Basque etymological dictionary~\cite{trask2008etymological}, and extract Basque words of unknown origins to have a better chance to match Iberian tokens. 

For all other known languages used for the closeness experiments, we use the Book of Genesis in these languages compiled by~\citet{christodouloupoulos2015massively} and take the most frequent stems. The number of stems is chosen to be roughly the same as the actual close relative, in order to remove any potential impact due to different vocabulary sizes. For instance, for the Gothic experiments in Figure~\ref{fig:ono_closeness}b, this number is set to be 600 since the PG vocabulary has 613 words.

\subsection{Training details}
\paragraph{Architecture} For the majority of our experiments, we use a dimensionality of 100 for each feature embedding, making the character embedding of size 700 (there are 7 feature groups). For ablation study without IPA embeddings, each character is directly represented by a vector of size 700 instead. To compare with previous work, we use the default setting from \texttt{Neurocipher} which has a hidden size of 250, and therefore for our model we use a feature embedding size of 35 , making it 245 for each character.

\paragraph{Hyperparameters}
We use SGD with a learning rate of 0.2 for all experiments.  Dropout with a rate of 0.5 is applied after the embedding layer. The length for matched spans $l$ in the range $[4, 10]$ for most experiments and $[3, 10]$ for Ugaritic. Other settings include $T=0.2, \lambda_{\text{cov}}=10.0, \lambda_{\text{cov}}=100.0$. We experimented with two annealing schedules for the insertion penalty $\alpha$: $\ln\alpha$ is annealed from $10.0$ to $3.5$ or from $0.0$ to $3.5$. These values are chosen based on our preliminary results, representing an extreme (10.0), a moderate (3.5) or a non-existent (0.0) penalty. Annealing last for 2000 steps, and the model is trained for an additional 1000 step afterwards. Five random runs are conducted for each setting and annealing schedule, and the best result is reported.

\end{document}